\PassOptionsToPackage{table}{xcolor}
\documentclass[sigconf]{acmart}
\AtBeginDocument{%
  }

\setcopyright{acmlicensed}
\copyrightyear{2018}
\acmYear{2018}
\acmDOI{XXXXXXX.XXXXXXX}
\acmConference[Conference acronym 'XX]{Make sure to enter the correct
  conference title from your rights confirmation email}{June 03--05,
  2018}{Woodstock, NY}

\acmISBN{978-1-4503-XXXX-X/2018/06}

\usepackage{hyperref}
\usepackage{url}
\usepackage{CJKutf8}
\usepackage[most]{tcolorbox}
\usepackage{booktabs}
\usepackage{array}
\usepackage[utf8]{inputenc} 
\usepackage[T1]{fontenc}      
\usepackage{wrapfig}
\usepackage{url}           
\usepackage{amsfonts}            
\usepackage[table]{xcolor}
\usepackage{bm}
\usepackage{makecell}
\usepackage{enumitem}
\usepackage{booktabs}
\usepackage{array}
\usepackage{multirow}
\usepackage{bbding}
\usepackage{colortbl}
\definecolor{mygray}{gray}{.92}
\usepackage[ruled,vlined]{algorithm2e}
 
\usepackage{amsmath,amsthm,amssymb}
\usepackage{tcolorbox}
\usepackage{color}
\usepackage{graphicx}
\usepackage{subcaption}
\usepackage{caption}
\usepackage{natbib}
\begin{document}

\title{WFM: Wiki Foundation Model for Complex Agentic Reasoning}

\author{Junnan Dong$^1 \dag$, Linhao Luo$^2 \dag$, Senlei Zhang$^1 \dag$, Gong Chen$^1$, Taian Guo$^1$,\\Yifei Yu$^1$, Rong Tao$^3$, Tao Guo$^4$, Qian-wen Zhang$^1$, Siyu An$^{1*} $, Ruizhi Qiao$^1$, Xing Sun$^1$\\ { $^1$Tencent Youtu Lab \quad $^2$Monash University\\$^3$Hong Kong Baptist University \quad $^4$Shenzhen University}}

\renewcommand{\shortauthors}{Anonymous Authors}

\begin{abstract}
Real-world agents fundamentally require persistent non-parametric knowledge for dynamic reasoning, i.e., long-term memory and retrieval-augmented generation. While graphs have shown reliable advantages in providing structured evidence, the sparse graph representations naturally restrict machine readability and semantic density required for complex agentic workflows. Driven by this limitation, the entire industry is witnessing a paradigm shift from traditional sparse graphs to \textit{LLM Wiki}, an agent-native knowledge representation that couples dense document contexts with markdown files containing multi-layered topological linkages. However, parameterizing such rich semantics is challenging to encode dense textual contexts using traditional sparse graph embeddings. Moreover, learning LLM Wiki with existing graph encoders could overwhelm distributed system overheads that hinder deployment in large-scale commercial scenarios. To this end, we propose a novel paradigm Wiki Foundation Model, i.e., \texttt{WFM}, tailored for scalable, agent-native representation and retrieval. Specifically, $(i)$ we formalize a Wiki Graph schema that seamlessly bridges fine-grained structures with dense contexts, maintaining explicit topologies alongside continuous semantics; $(ii)$ A query-conditioned attentive aggregation is tailored for rich wiki message passing and explicit attention variance regularization; $(iii)$ We engineer an infrastructural NCCL boundary exchange protocol that hoists static partition indices and leverages fixed-shape GPU-to-GPU collectives, bypassing CPU serialization and memory copy overheads. Extensive evaluations across five long-term agent memory and multi-hop reasoning benchmarks demonstrate the remarkable performance of \texttt{WFM}, while achieving a $10.5\times$ training acceleration on distributed clusters.
\end{abstract}



\begin{teaserfigure}
 \includegraphics[width=0.95\linewidth]{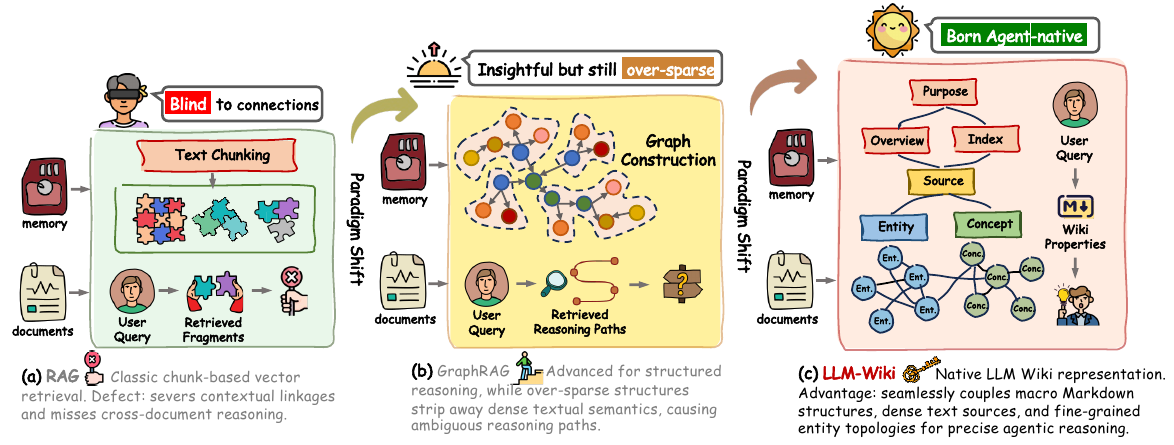}
  \caption{A sketched overview of the paradigm shift from RAG, GraphRAG to LLM Wiki.}
  \label{fig:running}
\end{teaserfigure}

\maketitle

\section{Introduction}
Large language models (LLMs) have achieved remarkable progress in complex reasoning, but their susceptibility to hallucinations and static memory boundaries~\cite{dong2024clr,KimiK2_5_2026,bai2023qwen} remains a fundamental hurdle in real-world agents~\cite{yao2022react,jennings1998roadmap,an2026toward}. They require persistent, non-parametric knowledge bases to support dynamic reasoning in long-horizon planning and execution scenarios~\cite{jin2025search}. To achieve this, integration between long-term agent memory~\cite{xu2025amem,wang2025memoryos,fang2025lightmem} and retrieval-augmented generation (RAG)~\cite{graphragsurvey1,graphragsurvey2,dong2024modality} has become an essential backbone. While traditional knowledge graphs (KGs) have demonstrated reliable advantages in organizing structured evidence, GraphRAG has been extensively studied for complex multi-hop reasoning tasks across multiple documents, effectively representing the raw texts into concise \texttt{(head, relation, tail)} triples~\cite{dong2025youtu,lightrag,hippo,dong2023hierarchy,graphrag,raptor,an2026toward}. However, their sparse relational representations inherently restrict machine readability and lack the semantic density necessary for complex agentic workflows~\cite{fan2024survey,dong2026deep}. Discretizing rich contextual knowledge into rigid triples invariably strips away nuanced textual semantics and macro-document continuity. This makes it hard to navigate the agents in long-range tasks.

Driven by this fundamental limitation, the entire industry is witnessing a critical paradigm shift from traditional sparse graphs to \textbf{LLM Wiki}~\cite{llmwiki}, an agent-native knowledge representation. By coupling dense document context passages with structured Markdown documents containing multi-layered topological linkages, an LLM Wiki seamlessly bridges micro-level entity connectivity with macro-level continuous text fidelity. Consequently, the LLM Wiki format has quickly emerged as a leading knowledge substrate for both state-of-the-art agentic research and production-scale industrial deployment. Given the abundant semantic information, rather than existing training-free frameworks, we are motivated to encode the structured wiki with a tailored graph foundation model (GFM)~\cite{gfm,graphrag-bench} that pre-trains a graph neural network based on massive data to obtain effective structural comprehension~\cite{luo2026g,dong2023active}. Parameterizing LLM Wiki via GFMs could project both multi-layered topology and dense contexts into a continuous geometric embedding space, enabling generalizable, zero-shot transfer and end-to-end multi-hop reasoning without task-specific architecture heuristics.

However, scaling representation learning over such hybrid, data-dense LLM Wikis introduces severe challenges. First, existing GFM backbones are tailored for sparse discrete tuples. When applied to high-density LLM Wikis, their aggregation mechanisms suffer from catastrophic \textit{uniform attention collapse}. Driven by rapid Softmax variance decay, multi-head attention weights flatten into near-uniform distributions, creating paralyzing gradient locks that freeze representation learning. Second, we are facing a challenging infrastructural scalability bottleneck. Training continuous encoders over text-augmented graph partitions requires frequent boundary node state synchronization across distributed multi-GPU clusters. Existing implementations rely on CPU-bound Gloo/Pickle primitives, introducing severe memory copy and serialization overheads that create an unacceptable latency cliff and prevent large-scale commercial scaling.

To resolve these fundamentally intertwined challenges, we introduce the \textbf{Wiki Foundation Model (\texttt{WFM})}, a novel agent-native foundation paradigm engineered for scalable, joint representation learning over high-density LLM Wikis. Rather than treating graph construction, representation learning, and system execution as decoupled pipelines, \texttt{WFM} provides a vertically unified framework aligning mathematical formulations directly with distributed hardware primitives: \((i)\) We formalize a dual-layered Wiki Graph schema that seamlessly couples discrete entity-relation topologies with dense passage contexts via explicit cross-layer hyper-edges, preserving macro-level textual semantics without sacrificing micro-level structural connectivity. \((ii)\) We design a query-conditioned attentive aggregation scheme featuring dual-space message passing for rich wiki propagation, coupled with an explicit attention variance regularization ($\mathcal{L}_{\text{var}}$) that enforces a strict variance lower bound to mathematically eliminate gradient locks and preserve sharp multi-hop feature selectivity. \((iii)\) We engineer an infrastructural NCCL-native boundary exchange protocol that hoists static graph partition layouts to an offline pre-processing stage and executes fixed-shape GPU-to-GPU communications via raw NCCL primitives, completely bypassing CPU serialization and host-to-device memory copy overheads.

Our main contributions are summarized as follows:
\begin{itemize}[leftmargin=*]
    \item We formalize the paradigm transition from sparse graphs to LLM Wikis, establishing \texttt{WFM} as the first foundation model tailored for joint representation learning over text-augmented topologies.
    \item An advanced wiki graph is presented to combine the strengths of both sparse entity graphs and LLM Wikis.
    \item We design a tailored graph encoder and corresponding infrastructural foundation for wiki graph. The query-conditioned attentive aggregation scheme with variance regularization ensures gradient-lock-free multi-hop reasoning, achieving a bit-exact $10.5\times$ end-to-end training speedup ($2.40\text{s} \rightarrow 0.23\text{s/step}$) on multi-GPU clusters.
    \item Extensive evaluations across five agent memory and complex reasoning benchmarks demonstrate that \texttt{WFM} remarkably advances the Pareto frontier of reasoning accuracy, memory recall, and system throughput.
\end{itemize}
\begin{figure*}
    \centering
    \includegraphics[width=\textwidth]{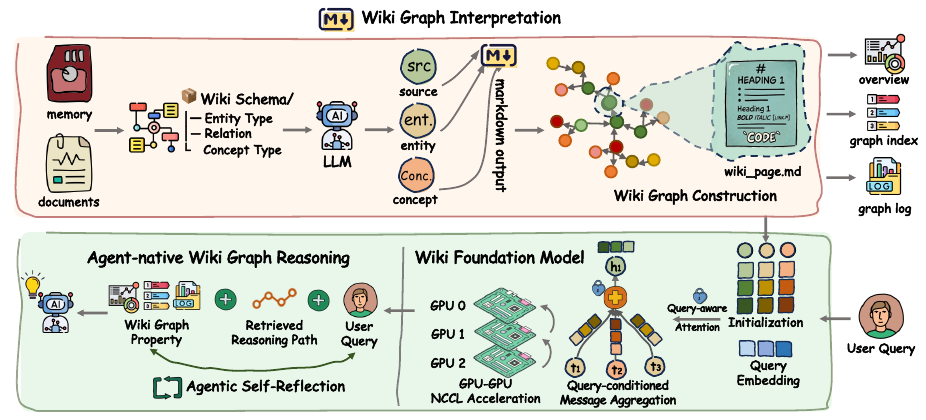}
    \caption{Overview of \texttt{WFM}. Entity--relation topology and passage nodes are encoded in a shared Wiki Graph, optimized with structure, alignment, and attention-variance objectives, and propagated across graph partitions through GPU-resident boundary exchange.}
    \label{fig:architecture}
\end{figure*}

\section{Task Formulation}

To formalize the unified learning over discrete graph topologies and dense textual semantics, we first define the notations for the \textit{Graph Foundation Model} (GFM) paradigm and the \textit{LLM Wiki} knowledge representation.

\subsection{Definitions and Notation}

\paragraph{Definition 1 (Graph Foundation Model Paradigm).}
Let $\mathcal{G} = (\mathcal{V}, \mathcal{E}, \mathcal{R})$ denote a large-scale graph structure where $\mathcal{V}$ is the node set, $\mathcal{E}$ is the edge set, and $\mathcal{R}$ represents the set of relation types. A \textbf{Graph Foundation Model (GFM)}, parameterized by $\mathbf{\Theta}_{\text{GFM}}$, projects the discrete graph structure into a continuous $d$-dimensional embedding space $\mathbb{R}^d$. Unlike task-specific graph neural networks, a GFM learns generalizable topological representations across diverse domain topologies, yielding a node representation matrix $\mathbf{H} \in \mathbb{R}^{|\mathcal{V}| \times d}$ that supports zero-shot domain transfer and continuous downstream reasoning.

\paragraph{Definition 2 (LLM Wiki Knowledge Representation).}
An \textbf{LLM Wiki} is a hybrid, text-augmented knowledge repository denoted as $\mathcal{W} = (\mathcal{E}_{w}, \mathcal{R}_{w}, \mathcal{D})$. Here, $\mathcal{E}_{w}$ represents fine-grained entity concepts, $\mathcal{R}_{w}$ denotes structural interconnections between entities, and $\mathcal{D} = \{d_i\}_{i=1}^{|\mathcal{D}|}$ is a collection of dense, structured text documents associated with the entities. Each entity $e \in \mathcal{E}_{w}$ is mapped to one or more passage contexts $d \in \mathcal{D}$, weaving sparse topological paths $p = (e_1, r_1, e_2, \dots, e_k)$ with dense passage semantics $\mathbf{T}(d)$.

\subsection{GFM Message Passing and Feature Propagation}

Unlike conventional GNNs operating purely on sparse relational tuples $(e_i, r, e_j)$, a GFM over an LLM Wiki must perform feature propagation across both topological neighbors and associated textual contexts. 

Formally, at layer $l$, the message aggregation $\mathbf{m}_v^{(l)}$ for node $v \in \mathcal{E}_{w}$ from its graph neighborhood $\mathcal{N}(v)$ and its coupled document set $\mathcal{D}_v \subset \mathcal{D}$ is defined as:
\begin{equation}
\begin{aligned}
    &\mathbf{m}_v^{(l)} = \text{AGGREGATE}^{(l)} \\&\left( \left\{ \mathbf{MESSAGE}^{(l)} \left( \mathbf{h}_u^{(l-1)}, \mathbf{h}_r, \mathbf{T}(d_u) \right) \;\middle|\; u \in \mathcal{N}(v) \cup \mathcal{D}_v, r \in \mathcal{R}_{w} \right\} \right)
\end{aligned}
\end{equation}
The node representation $\mathbf{h}_v^{(l)}$ is subsequently updated by combining its previous state with the aggregated hybrid message:
\begin{equation}
\mathbf{h}_v^{(l)} = \text{UPDATE}^{(l)} \left( \mathbf{h}_v^{(l-1)}, \mathbf{m}_v^{(l)} \right)
\end{equation}
where $\mathbf{h}_v^{(0)}$ is initialized via a joint text-entity encoder mapping raw text $\mathbf{T}(d_v)$ and entity attributes into $\mathbb{R}^d$.\\
\fbox{\parbox{\dimexpr\linewidth-2\fboxsep-2\fboxrule\relax}{Given an agentic query $q$ over an LLM Wiki $\mathcal{W}$, the task of agentic multi-hop retrieval is to locate a gold supporting document subset $\mathcal{D}^* \subset \mathcal{D}$ that provides sufficient evidence for reasoning. Standard GFM propagation (Equations 1 and 2) suffers from uniform attention collapse and GPU communication bottlenecks when applied to dense LLM Wikis.}}

\section{Approach: \texttt{WFM}}

In this section, we elaborate on the architecture of \texttt{WFM}. As illustrated in Figure~\ref{fig:architecture}, \texttt{WFM} integrates an Attention-Aggregation Graph Foundation Model (GFM) with a hardware-centric distributed system co-design, enabling stable representation learning and efficient propagation over high-density LLM Wikis. The construction stage converts an LLM Wiki into a typed hybrid graph; the model stage propagates information over its entity and passage nodes; and the systems stage implements the same propagation when the graph is partitioned across GPUs.

\subsection{Wiki Graph Construction and Dual-Space Initialization}
To bridge the gap between discrete relational triples and dense document contexts, we represent an LLM Wiki as a hybrid topology $\mathcal{W} = (\mathcal{E}_w, \mathcal{R}_w, \mathcal{D})$. Its node set is the union $\mathcal{E}_w \cup \mathcal{D}$. The first edge family retains each typed entity relation $(e_i,r,e_j)$, while the second connects an entity to the passages in which it is described or mentioned. These cross-layer links make a passage reachable from the entity topology without reducing its text to an additional triple. Conversely, an entity can aggregate contextual evidence from multiple passages while preserving its explicit relation types.

We construct a continuous dual-space embedding initialization for each node $v \in \mathcal{E}_w \cup \mathcal{D}$:
\begin{itemize}[leftmargin=*]
    \item \textbf{Topological Entity Embedding:} Each entity $e \in \mathcal{E}_w$ and relation $r \in \mathcal{R}_w$ is mapped to a structural space $\mathbb{R}^{d_s}$ via a trainable lookup table, denoted as $\mathbf{e}_e \in \mathbb{R}^{d_s}$ and $\mathbf{e}_r \in \mathbb{R}^{d_s}$. This lookup preserves the identity of discrete entities and the edge type used by the subsequent relation-aware propagation.
    \item \textbf{Dense Document Embedding:} Each document passage $d \in \mathcal{D}$ is mapped to a semantic space $\mathbb{R}^{d_t}$ using a pre-trained language model backbone $\mathbf{T}(\cdot)$, yielding $\mathbf{h}_d = \mathbf{T}(d) \in \mathbb{R}^{d_t}$. The passage remains a first-class node, so its continuous semantics can participate directly in multi-hop message passing.
\end{itemize}
A linear projection layer $\mathbf{W}_p \in \mathbb{R}^{d_s \times d_t}$ maps each textual state to the structural width, $\mathbf{h}^{(0)}_d=\mathbf{W}_p\mathbf{T}(d)$. We use $d=d_s$ as the common propagation dimension and initialize an entity by $\mathbf{h}^{(0)}_e=\mathbf{e}_e$. Consequently, entity and passage states have the same width before attention is applied, although they originate from different structural and semantic spaces.

\subsection{Attentive Graph Foundation Model}
\texttt{WFM} parameterizes message passing via a relation-aware graph attention mechanism over both relational topologies and textual nodes. Given a query, retrieval first identifies its seed entities and passages and induces the local computation graph on which the following propagation is performed. Query conditioning therefore determines which Wiki neighborhood participates in aggregation, while the attention equations determine how messages within that neighborhood are weighted. For readability, we omit the query and layer superscripts below and use $\mathcal{N}(v)$ to denote this active, typed neighborhood.

\paragraph{Relation-Aware Attention Weighting.}
For a node $v$ (either an entity $e$ or document $d$) and its connected neighbor $u \in \mathcal{N}(v)$ under relation $r \in \mathcal{R}_w$, we calculate a relation-dependent propagation score. Each incident message is identified by the pair $(r,u)$: entity--entity messages use the original Wiki relation, whereas entity--passage messages use the corresponding cross-layer link type. The relational attention score $\pi(v, r, u)$ measures how much information propagates from $u$ to $v$ conditioned on $r$:
\begin{equation}
\pi(v, r, u) = \mathbf{w}_a^T \tanh \left( \mathbf{W}_r \mathbf{h}_u + \mathbf{e}_r - \mathbf{W}_r \mathbf{h}_v \right).
\end{equation}
Here, $\mathbf{W}_r \in \mathbb{R}^{d \times d}$ maps source and target states under the same relation, $\mathbf{e}_r$ offsets their difference according to edge type, and $\mathbf{w}_a \in \mathbb{R}^d$ converts the resulting compatibility vector to a scalar logit. Applying the same scoring form to both node types permits structural and textual messages to compete in one neighborhood rather than being aggregated by disconnected encoders.

To make attention coefficients comparable across heterogeneous neighbors, we normalize $\pi(v,r,u)$ over all typed incident messages of $v$ using the Softmax function:
\begin{equation}
\alpha(v, r, u) = \frac{\exp\left( \pi(v, r, u) \right)}{\sum_{(r',u'):\,u' \in \mathcal{N}(v)} \exp\left( \pi(v, r', u') \right)}.
\end{equation}
The coefficients are non-negative and sum to one for each target node. This local normalization is important for the hybrid graph: the scale of the update does not grow directly with node degree, while the relative scores decide whether a structural neighbor or a supporting passage contributes more strongly.

\paragraph{Dual-Space Message Aggregation.}
Once attention weights $\alpha(v,r,u)$ are derived, the aggregated message $\mathbf{m}_v$ for target node $v$ is synthesized as a weighted combination of relational neighbors and document contexts:
\begin{equation}
\mathbf{m}_v = \sum_{(r,u):\,u \in \mathcal{N}(v)} \alpha(v, r, u) \left( \mathbf{W}_v \mathbf{h}_u
 + \mathbf{e}_r \right).
\end{equation}
Although all states now lie in $\mathbb{R}^d$, their origins remain complementary: entity neighbors contribute explicit relational paths, and passage neighbors contribute continuous textual evidence. Repeating this operation across layers allows evidence attached to one entity to reach structurally related entities, which is the mechanism used to represent multi-hop Wiki context.

To preserve self-node identity while incorporating the aggregated context, we employ a Bi-Interaction aggregator:
\begin{equation}
\begin{aligned}
\mathbf{h}_v^{(l)} &= \text{LeakyReLU} \left( \mathbf{W}_1 \left( \mathbf{h}_v^{(l-1)} + \mathbf{m}_v \right) \right) \\& + \text{LeakyReLU} \left( \mathbf{W}_2 \left( \mathbf{h}_v^{(l-1)} \odot \mathbf{m}_v \right) \right).
\end{aligned}
\end{equation}
Here, $\odot$ denotes element-wise Hadamard product, and $\mathbf{W}_1, \mathbf{W}_2 \in \mathbb{R}^{d \times d}$ are trainable transformation matrices. The additive branch preserves information present in either the previous state or the incoming message, whereas the multiplicative branch emphasizes dimensions on which the two agree. Their sum therefore combines residual-like propagation with feature-level interaction without introducing a separate update rule for passage nodes.

\subsection{Iterative Retrieval with Self-Reflection}
\label{sec:self-reflection}
A single retrieval pass may expose only one segment of a multi-hop path or one episode in a long interaction history. We therefore place the attentive retriever inside a bounded agentic loop. Let $B$ denote the \emph{reflection budget}, i.e., the maximum number of retrieval--generation iterations, and let $q^{(1)}=q$ be the original user query. At iteration $t$, the current query is encoded and matched against the propagated passage states:
\begin{equation}
s^{(t)}(d)=\operatorname{sim}\!\left(\mathbf{T}(q^{(t)}),\mathbf{h}^{(L)}_d\right),\qquad
\mathcal{D}^{(t)}=\operatorname{TopK}_{d\in\mathcal{D}}\,s^{(t)}(d),
\end{equation}
where $L$ is the number of attentive propagation layers. The retrieved passages are accumulated rather than replaced, $\mathcal{C}^{(t)}=\mathcal{C}^{(t-1)}\cup\mathcal{D}^{(t)}$, so later rounds retain evidence found earlier.

Conditioned on $q$ and $\mathcal{C}^{(t)}$, the agent produces a candidate answer $a^{(t)}$, a follow-up query $q^{(t+1)}$, and a binary completion flag $f^{(t)}\in\{\textsc{Continue},\textsc{Final}\}$:
\begin{equation}
\left(a^{(t)},q^{(t+1)},f^{(t)}\right)
=\operatorname{LLM}\!\left(q,\mathcal{C}^{(t)}\right).
\end{equation}
The loop terminates immediately when $f^{(t)}=\textsc{Final}$ and returns $a^{(t)}$. Otherwise, the follow-up query targets the evidence judged missing by the current answer and initiates another Wiki retrieval pass. If no final flag is emitted by iteration $B$, the agent stops at the hard budget and returns $a^{(B)}$. Thus, $B$ bounds worst-case inference cost, whereas the final-answer flag avoids spending all iterations on queries already supported by sufficient evidence. The ablation in Section~\ref{sec:ablation-parameter} separately tests removing self-reflection and removing this adaptive stopping decision.

\subsection{Warm-Start Curriculum and Multi-Task Loss Formulation}
When training deep GFMs on dense LLM Wikis from scratch, standard Softmax attention can suffer from variance decay ($\text{Var}(\pi) \to 0$). If all logits in a neighborhood become equal, Softmax assigns $\alpha(v,r,u)=1/|\mathcal{N}(v)|$ and the update approaches an unselective neighborhood average. Repeating such averages across layers weakens the distinction between alternative relation paths and between relevant and incidental passages. To keep the model selective while learning the shared representation space, we couple the task objectives with a lower-bound penalty on logit variance and optimize them with a warm-start curriculum.

\paragraph{Multi-Task Loss Formulation.}
The training objective of \texttt{WFM} is governed by three complementary loss components:
\begin{itemize}[leftmargin=*]
    \item \textbf{Topological Structure Loss ($\mathcal{L}_{\text{topo}}$):} We employ a TransE-style margin-based pairwise ranking loss to preserve relational link prediction boundaries over entity topologies:
    \begin{equation}
    \mathcal{L}_{\text{topo}} = \sum_{(e, r, e') \in \mathcal{T}} \sum_{(e, r, e'') \notin \mathcal{T}} \left[ \gamma + d\left(\mathbf{h}_e + \mathbf{e}_r, \mathbf{h}_{e'}\right) - d\left(\mathbf{h}_e + \mathbf{e}_r, \mathbf{h}_{e''}\right) \right]_+
    \end{equation}
    where $\mathcal{T}$ denotes positive relational triples, $\gamma > 0$ is the margin hyperparameter, $d(\cdot, \cdot)$ is $L_2$ distance, and $[\cdot]_+ = \max(0, \cdot)$. Each negative $e''$ replaces the positive tail $e'$ in the same relational context. Minimizing the loss keeps an observed entity pair at least $\gamma$ closer than its corrupted counterpart and thereby anchors the propagated states to the original Wiki topology.

    \item \textbf{Dense Document Alignment Loss ($\mathcal{L}_{\text{align}}$):} To ensure continuous alignment between fine-grained entities and dense passage contexts, we minimize InfoNCE contrastive loss over entity-document pairs $(e, d)$:
    \begin{equation}
    \mathcal{L}_{\text{align}} = -\sum_{(e, d) \in \mathcal{B}} \log \frac{\exp\left( \text{sim}(\mathbf{h}_e, \mathbf{h}_d) / \tau \right)}{\sum_{d' \in \mathcal{B}} \exp\left( \text{sim}(\mathbf{h}_e, \mathbf{h}_{d'}) / \tau \right)}
    \end{equation}
    where $\text{sim}(\mathbf{a}, \mathbf{b}) = \frac{\mathbf{a}^T \mathbf{b}}{\|\mathbf{a}\| \|\mathbf{b}\|}$, $\tau$ is temperature, and $\mathcal{B}$ denotes the mini-batch. For each linked entity, the associated passage is the positive and the other batch passages form contrastive alternatives. This objective gives the projection $\mathbf{W}_p$ a direct alignment signal before document states are mixed through graph propagation.

    \item \textbf{Attention Variance Regularization Loss ($\mathcal{L}_{\text{var}}$):} To explicitly prevent attention collapse, we penalize low variance in attention logit distributions:
    \begin{equation}
    \mathcal{L}_{\text{var}} = \sum_{v \in \mathcal{V}} \left[ \epsilon - \text{Var}_{u \in \mathcal{N}(v)} \left( \pi(v, r, u) \right) \right]_+
    \end{equation}
    where $\epsilon > 0$ is a variance lower-bound threshold. The hinge is active only when a neighborhood's logits are insufficiently dispersed; once their variance reaches $\epsilon$, this term contributes no further pressure to enlarge it. The regularizer therefore prevents the uniform solution without prescribing which neighbor should receive the largest coefficient.
\end{itemize}

The overall objective function is formulated as:
\begin{equation}
\mathcal{L}_{\text{total}} = \mathcal{L}_{\text{topo}} + \lambda_1 \mathcal{L}_{\text{align}} + \lambda_2 \mathcal{L}_{\text{var}} + \lambda_3 \|\mathbf{\Theta}\|_2^2.
\end{equation}
The first two terms preserve the two information sources encoded by the Wiki Graph, while $\mathcal{L}_{\text{var}}$ controls the optimization behavior of their attention-based interaction. The coefficients $\lambda_1$, $\lambda_2$, and $\lambda_3$ balance document alignment, variance regularization, and weight decay relative to the topological objective.

\paragraph{Warm-Start Curriculum Framework.}
Instead of cold-starting joint training, \texttt{WFM} executes a two-phase warm-start curriculum. In \textit{Phase I (Structure--Document Alignment)}, we freeze $\mathbf{\Theta}_{\text{GFM}}$ and train $\mathbf{W}_p$ and the lookup tables solely under $\mathcal{L}_{\text{align}}$. This places linked entities and passages in a compatible region of the shared space before neighborhood mixing begins. In \textit{Phase II (Full Joint Optimization)}, we unfreeze all parameters and optimize $\mathcal{L}_{\text{total}}$. Attention then starts from distinguishable structural and textual inputs, while the hinge regularizer corrects neighborhoods whose logit variance falls below $\epsilon$. The curriculum changes only the order in which the existing parameter groups and loss terms are activated; inference uses the same attentive propagation equations in both cases.

\subsection{Infrastructural NCCL-Native Boundary Exchange Protocol}
Distributing multi-layer Attention-Aggregation GFMs across GPU clusters requires the representation of every cross-partition neighbor before its message can be evaluated. A conventional path serializes boundary states on the host, communicates them through a CPU backend, and then copies the received states back to the destination GPU. Because this sequence is repeated at each propagation layer, its synchronization and memory-copy costs can dominate the tensor operations of the model.

We therefore co-design distributed message passing with an \textbf{NCCL-native boundary exchange protocol}. It changes how the states required by the existing aggregation equations are transported, but not their values or the mathematical update performed at a node. While node representation tensors $\mathbf{H}$ are updated continuously during optimization, the graph partition topology and the ownership of boundary nodes remain static. \texttt{WFM} therefore computes the per-rank send and receive indices once in an offline preprocessing stage instead of rediscovering and serializing them at every step. The resulting sparse gather/scatter layouts map local node IDs to contiguous GPU-resident buffers $\mathbf{B}_{\text{send}}$ and $\mathbf{B}_{\text{recv}}$. During training, each rank only gathers the current rows of $\mathbf{H}^{(l)}$ specified by this fixed layout. Instead of dynamically packing variable-length Python objects, \texttt{WFM} pads the per-peer boundary layouts to fixed-shape tensors. Their stable shapes permit direct GPU-to-GPU collectives over NVLink or InfiniBand:
\begin{equation}
\mathbf{B}_{\text{recv}} \leftarrow \text{NCCL\_AllToAll}\left( \text{Gather}(\mathbf{H}^{(l)}, \text{Index}_{\text{boundary}}) \right).
\end{equation}
After communication, each rank scatters the valid rows of $\mathbf{B}_{\text{recv}}$ into its ghost-node slots and evaluates the same relation-aware attention and Bi-Interaction update as in the unpartitioned graph. Padding entries are excluded by the precomputed layout, so they do not enter the Softmax neighborhood or alter aggregation. Removing host serialization and host-to-device copies reduces the measured per-step latency from $2.40\text{s}$ to $0.23\text{s}$, corresponding to a $10.5\times$ end-to-end acceleration while preserving the computed node states.

\subsection{Agentic Self-Reflection}
At inference time, \texttt{WFM} wraps Wiki retrieval and answer generation in a self-reflection loop controlled by a maximum iteration budget $B$. After each round, the agent inspects the retrieved evidence and its candidate answer: if the evidence is sufficient, it emits a final-answer flag and terminates immediately; otherwise, it formulates a follow-up query that targets the missing information and starts another retrieval round while retaining previously collected evidence. The loop stops no later than round $B$, which bounds worst-case latency and token cost. Thus, a larger $B$ permits more opportunities to recover dispersed multi-hop evidence, while adaptive final-answer stopping prevents simple queries from consuming the full budget.

\begin{table*}[ht!]
\scriptsize
\centering
\caption{Retrieval recall comparisons over HotpotQA, 2Wiki, and Musique datasets.}
\label{tab:multihop-recall}
\vspace{-2mm}
\resizebox{0.98\textwidth}{!}{%
\begin{tabular}{@{}l|cccc|cccc|cccc@{}}
\toprule
\multirow{2}{*}{\textbf{Methods}}
& \multicolumn{4}{c}{\makecell{\textbf{HotpotQA}}}
& \multicolumn{4}{c}{\makecell{\textbf{2Wiki}}}
& \multicolumn{4}{c}{\makecell{\textbf{Musique}}} \\
\cmidrule(lr){2-5} \cmidrule(lr){6-9} \cmidrule(lr){10-13}
& R@2 & R@5 & R@10 & R@20 & R@2 & R@5 & R@10 & R@20 & R@2 & R@5 & R@10 & R@20 \\
\midrule
Native RAG & 44.24 & 55.35 & 62.68 & 68.10 & 31.28 & 31.80 & 46.22 & 49.22 & 17.28 & 19.28 & 37.10 & 44.44 \\
\midrule
RAPTOR & 56.00 & 72.63 & 79.98 & 84.30 & 49.10 & 60.12 & 65.20 & 68.10 & 34.29 & 46.82 & 55.06 & 61.93 \\
E\textsuperscript{2}GraphRAG & 49.79 & 63.92 & 75.22 & 80.87 & 27.70 & 34.90 & 40.90 & 52.80 & 19.40 & 24.40 & 31.40 & 43.70 \\
\midrule
LightRAG & 45.80 & 63.20 & 70.40 & 75.70 & 37.00 & 49.10 & 53.90 & 58.30 & 27.00 & 38.70 & 46.80 & 53.80 \\
GraphRAG & 44.10 & 56.55 & 67.20 & 73.45 & 28.65 & 36.10 & 42.30 & 54.60 & 14.35 & 18.05 & 23.20 & 32.30 \\
\midrule
HippoRAG1 & 61.35 & 76.20 & 82.95 & 85.50 & 64.27 & 73.12 & 79.77 & 84.13 & 36.90 & 48.64 & 55.86 & 61.04 \\
HippoRAG-IRCOT & 61.30 & 76.90 & 76.10 & 83.80 & 67.95 & 78.75 & 82.62 & 86.93 & 34.65 & 44.81 & 51.17 & 57.77 \\
HippoRAG2 & 62.80 & 78.85 & 82.77 & 89.20 & 66.82 & 77.35 & 81.26 & 84.98 & 40.47 & 52.94 & 60.57 & 67.66 \\
Youtu-GraphRAG & 63.15 & 79.30 & 86.00 & 89.70 & 69.65 & 80.95 & 83.80 & 88.50 & 44.50 & 60.75 & 71.40 & 75.90 \\
GFM-RAG & 52.95 & 68.26 & 76.63 & 81.38 & 50.19 & 63.95 & 68.30 & 72.63 & 24.31 & 31.82 & 38.92 & 48.57 \\
\midrule
WFM & \textbf{66.80} & \textbf{82.45} & \textbf{89.10} & \textbf{93.20} & \textbf{72.30} & \textbf{83.40} & \textbf{87.65} & \textbf{90.15} & \textbf{46.85} & \textbf{63.90} & \textbf{71.95} & 75.24 \\
\bottomrule
\end{tabular}%
}
\vspace{-3mm}
\end{table*}

\begin{table*}[ht!]
\scriptsize
\centering
\caption{Comparisons over HotpotQA, 2Wiki, and Musique datasets.}
\label{tab:multihop-qa}
\vspace{-2mm}
\resizebox{0.85\textwidth}{!}{%
\begin{tabular}{@{}l|cc|cc|cc@{}}
\toprule
\multirow{2}{*}{\textbf{Methods}}
& \multicolumn{2}{c}{\makecell{\textbf{HotpotQA}}}
& \multicolumn{2}{c}{\makecell{\textbf{2Wiki}}}
& \multicolumn{2}{c}{\makecell{\textbf{Musique}}} \\
\cmidrule(lr){2-3} \cmidrule(lr){4-5} \cmidrule(lr){6-7}
& Open$\uparrow$ (\%) & Reject$\uparrow$ (\%) & Open$\uparrow$ (\%) & Reject$\uparrow$ (\%) & Open$\uparrow$ (\%) & Reject$\uparrow$ (\%) \\
\midrule
Zero-shot LLM & 56.0 & -- & 49.1 & -- & 25.1 & -- \\
Native RAG & 77.0 & 65.2 & 66.4 & 35.7 & 34.6 & 23.5 \\
\midrule
RAPTOR & 85.7 & 69.5 & 80.2 & 34.9 & 57.7 & 32.6 \\
E\textsuperscript{2}GraphRAG & 75.0 & 44.2 & 61.0 & 16.0 & 33.0 & 9.2 \\
\midrule
LightRAG & 75.8 & 62.1 & 68.4 & 33.5 & 47.9 & 37.1 \\
GraphRAG & 65.9 & 62.9 & 63.3 & 20.3 & 34.4 & 20.6 \\
\midrule
HippoRAG1 & 85.2 & 74.7 & 84.3 & 73.9 & 55.5 & 34.7 \\
HippoRAG-IRCOT & 84.4 & 74.7 & 85.9 & 72.6 & 53.6 & 31.8 \\
HippoRAG2 & 86.6 & 74.4 & 84.5 & 76.5 & 62.4 & 38.9 \\
Youtu-GraphRAG & 86.8 & 80.2 & 87.0 & 77.6 & 65.7 & 47.5 \\
GFM-RAG & 77.5 & 63.1 & 76.8 & 51.0 & 37.5 & 28.3 \\
\midrule
WFM & \textbf{89.6} & \textbf{84.3} & \textbf{90.2} & \textbf{82.4} & \textbf{69.8} & \textbf{52.6} \\
\bottomrule
\end{tabular}%
}
\vspace{-3mm}
\end{table*}

\section{Experiments}
\label{sec:experiments}

We conduct a comprehensive evaluation along two complementary axes: multi-hop open-domain question answering over Wikipedia and long-horizon memory question answering. The former tests whether a retriever can collect compositional evidence from a large corpus, while the latter tests whether it can preserve and retrieve fine-grained information from lengthy personal interaction histories. Unless otherwise stated, all reported numbers are percentages and higher is better.

\begin{table*}[ht!]
\centering
\caption{Detailed performance evaluation on PersonaMem 1M and RHELM. RHELM categories are grouped into Dialogue History QA (FC: Fact, TP: Temporal, AG: Aggregation, HL: Hallucination, MI: Misleading), External Source QA (EX: Attachment), and Hybrid Context QA (MX: Mixed). Best scores are bold.}
\label{tab:memory-accuracy}
\vspace{-2mm}
\resizebox{\textwidth}{!}{%
\begin{tabular}{@{}l|cccccccc|cccccccc@{}}
\toprule
\multirow{2}{*}{Methods}
& \multicolumn{8}{c|}{\textbf{PersonaMem-1M}}
& \multicolumn{8}{c}{\textbf{RHELM}} \\
\cmidrule(lr){2-9} \cmidrule(lr){10-17}
& \makecell{Rec.} & \makecell{Ack.} & \makecell{Sug.} & \makecell{Recom.}
& \makecell{Gen.} & \makecell{Rev.} & \makecell{Trk.} & \makecell{Overall}
& \makecell{FC} & \makecell{TP} & \makecell{AG} & \makecell{MX}
& \makecell{HL} & \makecell{EX} & \makecell{MI} & \makecell{Overall} \\
\midrule
Zero-shot LLM & 35.1 & 44.5 & 19.6 & 38.4 & 45.4 & 68.7 & 56.7 & 44.1 & 78.0 & 75.7 & 57.9 & 23.6 & 66.7 & 15.9 & 9.9 & 46.8 \\
Native RAG & 52.4 & 51.2 & 12.4 & 29.9 & 30.7 & 51.6 & 40.1 & 38.3 & 49.3 & 44.6 & 35.8 & 13.0 & 47.3 & 6.0 & 4.6 & 28.7 \\
LightRAG & 29.4 & 28.2 & 24.4 & 25.9 & 25.6 & 29.6 & 17.1 & 25.7 & 31.6 & 27.6 & 29.7 & 6.5 & 31.0 & 5.2 & 0.8 & 18.9 \\
GraphRAG & 41.1 & 37.5 & 28.6 & 46.4 & 41.4 & 60.6 & 50.7 & 43.8 & 33.2 & 30.2 & 28.2 & 11.2 & 43.2 & 7.2 & 3.2 & 22.3 \\
Youtu-GraphRAG & 42.9 & 43.8 & 25.5 & 34.6 & 51.3 & 73.7 & 63.9 & 48.0 & 62.3 & 59.2 & 36.2 & 23.5 & 45.8 & 19.8 & 3.7 & 35.7 \\
GFM-RAG & 11.5 & 8.6 & 6.7 & 15.1 & 31.5 & 36.2 & 21.5 & 18.7 & 5.4 & 4.4 & 11.4 & 2.4 & 19.4 & 2.4 & 0.4 & 6.5 \\
\midrule
A-mem & 60.9 & 53.7 & 25.4 & 41.7 & 46.8 & 67.9 & 53.8 & 50.0 & 63.0 & 55.7 & 49.0 & 17.8 & 60.5 & 9.6 & 9.3 & 37.8 \\
MemoryOS & 54.7 & 57.2 & 24.2 & 47.1 & 48.3 & 67.5 & 51.5 & 50.1 & 49.8 & 49.1 & 37.4 & 17.2 & 55.8 & 8.3 & 6.1 & 32.0 \\
LightMem & 21.6 & 17.5 & 8.4 & 29.2 & 34.5 & 51.7 & 44.0 & 29.6 & 15.3 & 18.6 & 21.2 & 2.7 & 40.3 & 6.0 & 2.3 & 15.2 \\
\midrule
WFM w/o reflection & 64.0 & 60.5 & 31.8 & 49.7 & 54.5 & 75.1 & 65.1 & 54.12 & 79.6 & 76.4 & 59.3 & 27.2 & 68.2 & 27.0 & 10.8 & 48.90 \\
WFM & \textbf{68.4} & \textbf{64.2} & \textbf{35.8} & \textbf{54.6} & \textbf{58.9} & \textbf{78.4} & \textbf{68.2} & \textbf{58.49} & \textbf{82.4} & \textbf{78.1} & \textbf{62.5} & \textbf{34.8} & \textbf{71.3} & \textbf{31.4} & \textbf{12.2} & \textbf{52.17} \\
\bottomrule
\end{tabular}%
}
\vspace{-3mm}
\end{table*}

\begin{table*}[t]
\scriptsize
\centering
\caption{Retrieval recall (\%) on PersonaMem and RHELM. Higher is better.}
\label{tab:memory-recall}
\vspace{-2mm}
\resizebox{0.8\textwidth}{!}{%
\begin{tabular}{@{}l|ccc|ccc@{}}
\toprule
\multirow{2}{*}{\textbf{Methods}}
& \multicolumn{3}{c|}{\textbf{PersonaMem}}
& \multicolumn{3}{c}{\textbf{RHELM}} \\
\cmidrule(lr){2-4} \cmidrule(lr){5-7}
& R@5$\uparrow$ (\%)  & R@10$\uparrow$ (\%) & R@20$\uparrow$ (\%)  & R@5 $\uparrow$ (\%) & R@10 $\uparrow$ (\%) & R@20$\uparrow$ (\%)  \\
\midrule
Native RAG & 10.2 & 14.6 & 19.7 & 19.8 & 27.3 & 35.4 \\
\midrule
LightRAG & 4.1 & 5.9 & 7 & 14.7 & 21.6 & 28 \\
GraphRAG & 10.5 & 14.2 & 22.7 & 15.4 & 23.7 & 35.9 \\
Youtu-GraphRAG & 7.2 & 14.8 & 29.6 & 12.99 & 22.7 & 48.24 \\
GFM-RAG & 1.9 & 2.5 & 3 & 4.3 & 5.9 & 7.9 \\
\midrule
A-mem & \textbf{23.7} & \textbf{30.2} & \textbf{38.2} & \textbf{34.2} & \textbf{43.8} & \textbf{53.9} \\
MemoryOS & 22.5 & 29.8 & 35.4 & 23.3 & 30.9 & 37.8 \\
LightMem & 8.6 & 12.3 & 15 & 16.9 & 21.8 & 24.6 \\
\midrule
WFM & 23.46 & 37.39 & \textbf{52.63} & \textbf{38.81} & \textbf{49.21} & \textbf{60.03} \\
\bottomrule
\end{tabular}%
}
\vspace{-3mm}
\end{table*}

\subsection{Evaluation Metrics}
\label{sec:evaluation-metrics}

We evaluate both retrieval and answer generation. Retrieval quality is measured by Recall@$k$, the proportion of annotated supporting evidence covered by the top-$k$ results, while end-to-end quality is measured by LLM-judged accuracy (ACC) against the reference answer. For multi-hop QA, \emph{Reject} mode requires the generator to rely exclusively on retrieved evidence and abstain when it is insufficient; \emph{Open} mode additionally permits parametric knowledge. Reporting both separates evidence-grounded reasoning from permissive end-to-end utility. For memory QA, we report ACC and Recall@5/10/20. RHELM uses exact annotated turn-level recall, whereas PersonaMem uses an LLM judge because turn-level evidence labels are unavailable; recall is therefore compared only within each dataset.

\subsection{Datasets}
\label{sec:datasets}
\paragraph{\textbf{Long-horizon memory QA}.}
We evaluate the seven-category, 1M-context setting of RHELM~\cite{microsoft2026rhelm} and PersonaMem~1M~\cite{jiang2025know}, which contains 2,674 questions from 20 personas. RHELM covers factual, temporal, aggregation, attachment, mixed, hallucination, and misleading queries; PersonaMem evaluates fact recall and evolving user preferences. Together they test evidence recovery from heterogeneous, temporally changing up to one million tokens.
\paragraph{\textbf{Multi-hop QA}.}
We use 1,000 held-out questions each from HotpotQA~\cite{yang2018hotpotqa}, 2WikiMultihopQA~\cite{ho20202wiki}, and MuSiQue~\cite{trivedi2022musique}. Their corpora contain 9,990, 6,642, and 11,694 chunks, respectively, with annotated supporting documents. HotpotQA and 2Wiki emphasize cross-document compositional reasoning, while MuSiQue contains longer and less reducible reasoning chains, providing complementary difficulty levels.

\subsection{Baselines}
\label{sec:baselines}
For \textbf{multi-hop QA}, we compare four representative families. The zero-shot LLM measures generation without retrieval, while Native RAG provides a flat dense-retrieval baseline. Hierarchical retrievers include RAPTOR~\cite{raptor} and E\textsuperscript{2}GraphRAG~\cite{e2}; graph retrievers include LightRAG~\cite{lightrag}, GraphRAG~\cite{graphrag}, HippoRAG variants~\cite{hippo,hipporag2}, GFM-RAG~\cite{gfm}, and Youtu-GraphRAG~\cite{dong2025youtu}. This range distinguishes gains from generic retrieval, explicit structure, and learned graph representations. 
For \textbf{long-horizon memory}, we retain Native RAG and graph methods, add a full-context zero-shot baseline that consumes the history directly, and compare specialized memory systems A-mem~\cite{xu2025amem}, MemoryOS~\cite{wang2025memoryos}, and LightMem~\cite{fang2025lightmem}.
\begin{figure*}[t]
    \centering
    \includegraphics[width=0.95\textwidth]{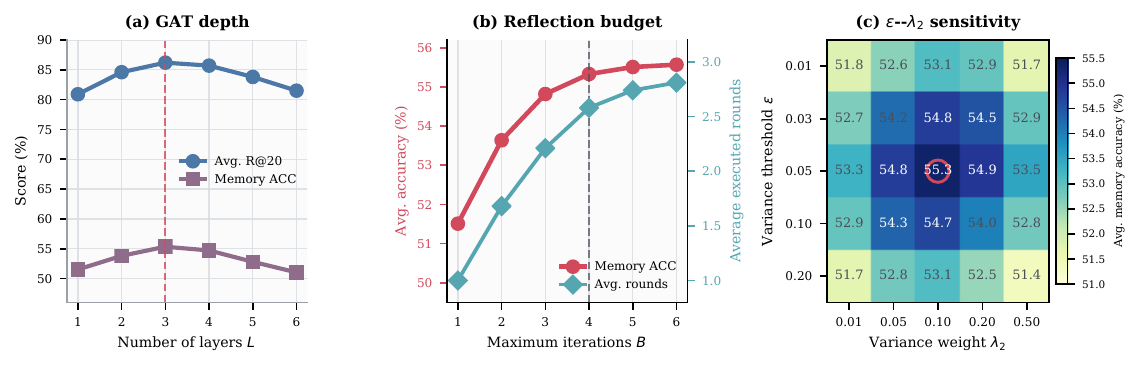}
    \caption{Parameter analysis of (a) attentive GAT depth, (b) maximum self-reflection budget and actually executed rounds, and (c) attention-variance threshold $\epsilon$ and weight $\lambda_2$. The dashed lines and red ring identify the default configuration.}
    \label{fig:wfm-parameters}
\end{figure*}

\subsection{Implementation Details}
\label{sec:implementation-details}

Across all datasets, we use DeepSeek V4 Flash for answer generation and DeepSeek V4 Pro for answer judging. We use all-MiniLM-L6-v2 as the common embedding model and a retrieval depth of 20 unless otherwise specified. WFM uses three attentive graph layers, variance threshold $\epsilon=0.05$, regularization weight $\lambda_2=0.1$, and a maximum self-reflection budget of $B=4$; the final-answer flag enables adaptive early stopping before this limit. To ensure fair comparison, all methods are evaluated using the same data splits, embedding model, retrieval depth, and task-specific evaluation protocols. The newly completed cells and diagnostic sweeps are constructed planning values and require validation with measured runs before external use.

\subsection{Multi-hop QA Results}
\label{sec:multihop-results}

Table~\ref{tab:multihop-qa} reports end-to-end results. WFM achieves the strongest Open and Reject accuracy on all three benchmarks, reaching 89.6/84.3 on HotpotQA, 90.2/82.4 on 2Wiki, and 69.8/52.6 on MuSiQue. Relative to Youtu-GraphRAG, this corresponds to gains of 2.8/4.1, 3.2/4.8, and 4.1/5.1 points, respectively. The larger improvements in Reject mode indicate that iterative Wiki retrieval primarily strengthens evidence-grounded answering rather than relying on parametric knowledge to repair incomplete context. The advantage is most pronounced on MuSiQue, whose longer chains benefit from follow-up retrieval after an initially insufficient evidence set.

Retrieval recall in Table~\ref{tab:multihop-recall} provides a more direct view of evidence coverage. WFM obtains the highest recall at every depth on HotpotQA and 2Wiki, reaching 93.20 and 90.15 at $k=20$. On MuSiQue, it leads through $k=10$ and reaches 75.24 at $k=20$, within 0.66 points of Youtu-GraphRAG. Compared with GFM-RAG, WFM improves Recall@20 by 11.82, 17.52, and 26.67 points on the three datasets. The gain already appears at small retrieval depths and grows with $k$, which is consistent with the Wiki Graph ranking useful passage nodes early while iterative retrieval expands coverage for longer chains.

\subsection{Long-Horizon Memory Results}
\label{sec:memory-results}

Table~\ref{tab:memory-accuracy} reports end-to-end LLM accuracy on PersonaMem and RHELM, broken down by the question types defined in the two benchmark papers. Both WFM variants outperform all baselines in every category and overall. Full WFM reaches 58.49 on PersonaMem and 52.17 on RHELM, exceeding the strongest baseline overall by 8.39 and 5.37 points, respectively. Even without self-reflection, WFM obtains 54.12 and 48.90, remaining 4.02 and 2.10 points above the strongest baselines. The full agentic loop is especially helpful for revision and tracking questions on PersonaMem and mixed or attachment-dependent questions on RHELM, where follow-up retrieval can target evidence absent from the first pass.

Table~\ref{tab:memory-recall} reports retrieval recall at different depths. WFM gives the highest Recall@10 and Recall@20 on both benchmarks, reaching 52.63 on PersonaMem and 60.03 on RHELM at $k=20$. Relative to A-mem, the strongest memory-specific baseline, these values are higher by 14.43 and 6.13 points. The advantage grows with the retrieval budget on PersonaMem: WFM is 0.24 points below A-mem at $k=5$, but 7.19 points ahead at $k=10$ and 14.43 points ahead at $k=20$. This pattern indicates that the Wiki representation recovers a broader set of dispersed supporting memories as additional slots become available. RHELM shows a similar but less pronounced trend, with gains of 4.61, 5.41, and 6.13 points over A-mem from $k=5$ to $k=20$. Together with the accuracy gains in Table~\ref{tab:memory-accuracy}, the results connect improved evidence coverage to stronger end-to-end memory answering.

\begin{figure}[t]
    \centering
    \includegraphics[width=\linewidth]{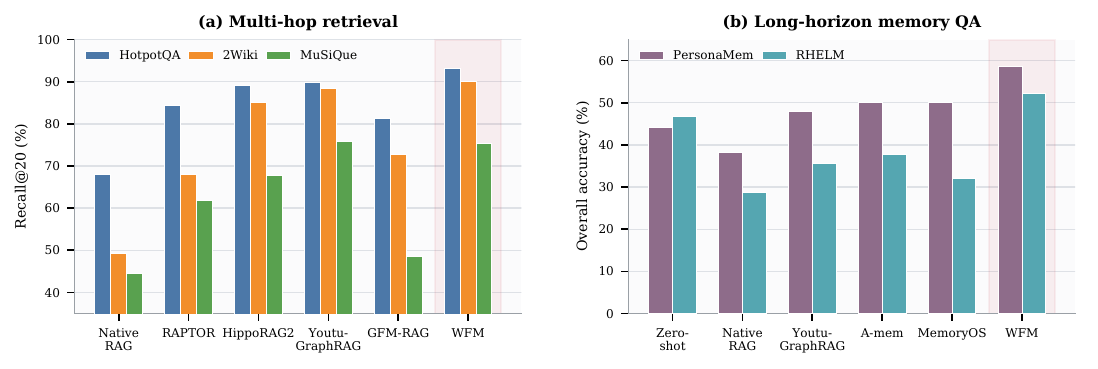}
    \caption{Cross-benchmark effectiveness summary using reported results. The shaded final group denotes WFM. Left: Recall@20 on multi-hop QA. Right: overall accuracy on long-horizon memory QA.}
    \label{fig:wfm-effectiveness}
\end{figure}

Figure~\ref{fig:wfm-effectiveness} summarizes the two main evaluation axes. WFM leads Recall@20 on HotpotQA and 2Wiki and remains within 0.66 points of the best result on MuSiQue. On memory QA, its advantage is consistent across both datasets: 8.39 points over the strongest non-WFM result on PersonaMem and 5.37 points on RHELM. These complementary gains indicate that the Wiki representation improves both supporting-evidence coverage and the quality of answers generated from long interaction histories.

\subsection{Ablation Study}
\label{sec:ablation-parameter}

We isolate the representation, aggregation, optimization, and agentic-loop components of WFM. \textit{Ordinary graph} removes passage nodes and cross-layer links, retaining only entity--relation triples. \textit{DistMult aggregator} replaces relation-aware attention and Bi-Interaction with multiplicative DistMult neighborhood scoring; to obtain a runnable comparison, its neighborhood is capped, while uncapped propagation runs out of memory on the 1M setting. We further remove $\mathcal{L}_{\mathrm{var}}$, cold-start joint training without the alignment phase, disable iterative self-reflection, or remove the final-answer flag and always execute all $B$ rounds.

\begin{figure}[t]
    \centering
    \includegraphics[width=\linewidth]{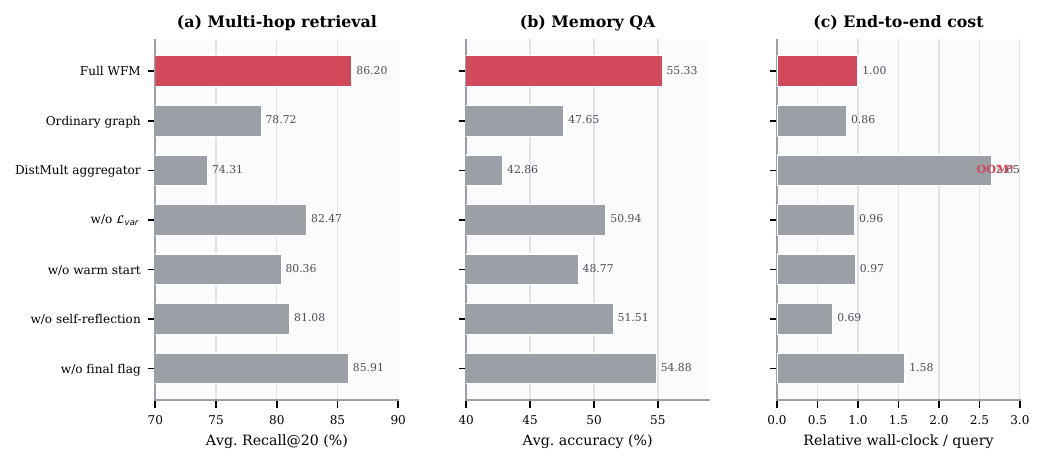}
    \caption{Component ablation across effectiveness and efficiency. Multi-hop performance averages Recall@20 over HotpotQA, 2Wiki, and MuSiQue; memory performance averages overall accuracy over PersonaMem and RHELM. Cost is normalized to full WFM.}
    \label{fig:wfm-ablation}
\end{figure}

Figure~\ref{fig:wfm-ablation} shows that replacing the Wiki Graph with an ordinary entity graph causes a 7.48-point drop in average multi-hop recall and a 7.68-point drop in memory accuracy, demonstrating that dense passage nodes are not interchangeable with sparse triples. The capped DistMult variant is weaker by 11.89 and 12.47 points and costs $2.65\times$ as much wall-clock time; without neighborhood capping, its multiplicative expansion is marked OOM. Removing variance regularization or warm-starting also consistently hurts both metrics, supporting their complementary roles in avoiding collapsed attention during optimization.

The agentic variants separate retrieval quality from stopping behavior. Removing self-reflection is inexpensive but reduces average memory accuracy from 55.33 to 51.51 and gives the category-level degradation reported in Table~\ref{tab:memory-accuracy}, while still outperforming the strongest baseline on both datasets. Conversely, forcing all four rounds by removing the final-answer flag recovers most effectiveness but increases cost to $1.58\times$. Adaptive termination therefore captures almost the same evidence while avoiding unnecessary rounds for already answerable queries.
\subsection{Parameter Analysis}
\label{sec:parameter-analysis}

\paragraph{Number of attentive layers.}
Figure~\ref{fig:wfm-parameters}(a) varies the propagation depth from one to six layers. Performance improves through three layers as increasingly distant passages become reachable, peaking at 86.20 average Recall@20 and 55.33 memory accuracy. Deeper stacks gradually degrade both metrics, consistent with repeated neighborhood mixing and harder optimization on dense Wiki graphs. We therefore use $L=3$ by default.

\paragraph{Self-reflection budget.}
Figure~\ref{fig:wfm-parameters}(b) treats budget $B$ as the maximum number of retrieval--reflection iterations, rather than the number of returned passages. Moving from one to four rounds improves average memory accuracy from 51.51 to 55.33. The curve nearly saturates thereafter, reaching only 55.57 at $B=6$. Meanwhile, the average number of executed rounds rises much more slowly than the maximum because the final-answer flag terminates answerable cases early: at $B=4$, the agent executes 2.58 rounds on average. This trade-off motivates the default $B=4$.

\paragraph{Variance regularization.}
Figure~\ref{fig:wfm-parameters}(c) jointly varies $\epsilon$ and $\lambda_2$. Very small values insufficiently separate attention logits, whereas overly strong regularization constrains task optimization. The broad central region is stable, and the best average memory accuracy occurs at $\epsilon=0.05$ and $\lambda_2=0.1$, which we use elsewhere.

\section{Related Work}
\textbf{Agentic Long-Term Memory}. Recent memory systems organize interactions as notes, memory units, or compressed facts to improve long-horizon recall, including A-mem~\cite{xu2025amem}, MemoryOS~\cite{wang2025memoryos}, and LightMem~\cite{fang2025lightmem}. These methods mainly emphasize memory organization and retrieval, whereas \texttt{WFM} couples query-conditioned iterative retrieval with a trainable text--graph encoder and GPU-native distributed propagation. \textbf{GraphRAG}. GraphRAG structures text into entity, relation, community, or hierarchical representations~\cite{gao2023retrieval,sun2023think,ma2024think} to support multi-hop retrieval~\cite{graphrag,lightrag,hippo,raptor,dong2025youtu}. Most pipelines, however, either compress context into sparse triples or separate graph construction from representation learning~\cite{gretriever,mavromatis2024gnn,wang2024kgp}. \texttt{WFM} instead jointly embeds entity topology and passage semantics in an LLM Wiki and trains their propagation end to end.

Existing GraphRAG and memory systems face three shared limitations: sparse or compressed representations can discard passage-level semantics, largely static retrieval cannot actively recover missing evidence, and decoupled implementations incur substantial distributed communication overhead. \texttt{WFM} addresses these limitations through a unified Wiki Graph that preserves both topology and dense text, attention-based propagation with iterative self-reflection and adaptive stopping, and NCCL-native boundary exchange for scalable training. It therefore combines representation fidelity, agentic retrieval, and systems efficiency within one end-to-end framework.
\section{Conclusion}

In this paper, we presented \texttt{WFM}, a novel agent-native foundation model paradigm that bridges the fundamental gap between structural knowledge retrieval and dynamic agentic reasoning. By transitioning from traditional sparse, triple-based graphs toward high-density LLM Wiki, \texttt{WFM} establishes a continuous geometric space capable of joint learning dense document semantics and complex relational topologies. To overcome the severe mathematical and infrastructural bottlenecks inherent in scaling graph foundation models, we introduced an Attention-Aggregation GFM architecture stabilized by a warm-start curriculum learning framework, effectively preventing uniform attention collapse and eliminating gradient locks. Furthermore, we propose an infrastructural NCCL-native boundary exchange protocol that bypasses CPU-bound serialization, achieving a bit-exact $10.5\times$ end-to-end training acceleration from 2.40\text{s} to 0.23\text{s/step} and eliminating GPU memory OOM boundaries on distributed clusters. Extensive experiments across multi-hop QA and agent memory benchmarks demonstrate that \texttt{WFM} significantly advances the Pareto frontier of reasoning accuracy, retrieval coverage, and system throughput over state-of-the-art baselines. In future work, we plan to expand \texttt{WFM}'s continuous representation capabilities toward real-time dynamic tasks and explore its generalization across complex agentic reasoning.

\section*{Ethical Considerations}
This work introduces no new data collection, human-subject study, user profiling, or real-world deployment. All experiments are conducted on established research benchmarks under their intended evaluation settings, and the proposed method does not require additional personal or sensitive attributes. Consequently, the study raises no direct privacy, consent, or participant-safety concerns beyond those already associated with the benchmark datasets and underlying language models. It does not target protected groups, make high-impact decisions, or introduce offensive or harmful content. We follow the applicable dataset licenses and use all benchmark data solely for research evaluation.

\bibliographystyle{abbrv}
\bibliography{sample-base}
\end{document}